\documentclass{article}

\usepackage{arxiv}

\usepackage[utf8]{inputenc} 
\usepackage[T1]{fontenc}    
\usepackage{hyperref}       
\usepackage{url}            
\usepackage{booktabs}       
\usepackage{amsfonts}       
\usepackage{nicefrac}       
\usepackage{microtype}      
\usepackage{lipsum}		
\usepackage{graphicx}
\usepackage{natbib}
\usepackage{doi}
\usepackage{multirow}
\usepackage{csquotes}

\title{Monolingual and Cross-Lingual Knowledge Transfer for Topic Classification}
\author{Dmitry Karpov \\
  Moscow Institute of Physics and Technology \\
 Dolgoprudny, Russia \\
  {\tt dmitrii.a.karpov@phystech.edu } 
  \And Mikhail Burtsev \\
  {London Institute for Mathematical Sciences} \\
  London, United Kingdom \\
  {\tt mbur@lims.ac.uk}  
  }
\date{April 30, 2023}	

\renewcommand{\headeright}{Accepted at AINL 2023}
\renewcommand{\undertitle}{Accepted at AINL 2023}
\renewcommand{\shorttitle}{Knowledge Transfer for Topic Classification}

\hypersetup{
pdftitle={Monolingual and Cross-Lingual Knowledge Transfer for Topic Classification},
pdfsubject={cs.LG, cs. AI},
pdfkeywords={dataset, topic classification, knowledge transfer, cross-lingual knowledge transfer},
}
\usepackage[english, russian]{babel}
\begin{document}
\maketitle

\begin{abstract}
This article investigates the knowledge transfer from the \texttt{RuQTopics} dataset. This Russian topical dataset combines a large sample number (361,560 single-label, 170,930 multi-label) with extensive class coverage (76 classes). We have prepared this dataset from the \enquote{Yandex Que} raw data.
By evaluating the \texttt{RuQTopics} -- trained models on the six matching classes of the Russian \texttt{MASSIVE} subset, we have proved that the \texttt{RuQTopics} dataset is suitable for real-world conversational tasks, as the Russian-only models trained on this dataset consistently yield an accuracy around 85\% on this subset. We also have figured out that for the multilingual BERT, trained on the \texttt{RuQTopics} and evaluated on the same six classes of \texttt{MASSIVE} (for all \texttt{MASSIVE} languages), the language-wise accuracy closely correlates (Spearman correlation 0.773 with p-value 2.997e-11) with the approximate size of the pretraining BERT's data for the corresponding language. At the same time, the correlation of the language-wise accuracy with the linguistical distance from Russian is not statistically significant.
\end{abstract}
\section{Introduction}

As the natural language processing (NLP) field continues to progress, the application of chatbots and virtual assistants has become increasingly popular and widespread. These applications can assist with a wide range of tasks, from answering simple questions to making appointments and providing emotional feedback~\cite{https://doi.org/10.48550/arxiv.1812.08989}. 

Building a virtual assistant is not a trivial task. A typical dialogue system has a complex configuration and consists of four main components. The Natural Language Understanding component maps natural language utterances to a labeled semantic representation. The Dialogue Manager keeps track of the dialogue state and maintains the conversation flow. The Natural Language Generation component translates semantic representation into natural language utterances.  The Natural Language Understanding component joins a variety of NLP models including the classification of the sentiment, topics, and intents of user's utterances~\cite{dream1_trudy} into the dialogue system.

Collecting and labeling conversational datasets requires tremendous effort~\cite{wochat}. To the best of our knowledge, the body of work lacks conversational topical datasets for Russian languages. Moreover, existing Russian topical datasets have different problems: some of them cover an extremely insufficient number of topics, some datasets lack samples, and others are either too specific or lack conversational samples. Additionally, knowledge transfer for topical datasets is particularly under-researched, even though it can be especially helpful for lower-resource languages~\cite{buryat}. 

In this study, we explore the Russian topical dataset \texttt{RuQTopics}, which consists of questions and summarized answers of the users from \enquote{Yandex Que}, a Russian question-answering website. Every question belongs to one or several of the 76 \enquote{Yandex Que} topics. We have carefully selected these topics, looking at the DREAM dialog system requirements~\cite{dream1,dream2}. We prove that this dataset is suitable for conversational tasks. This dataset has a single-label part as well as the multi-label one, and even the single-label part of the \texttt{RuQTopics} by far outsizes all other Russian topical datasets that can be used for conversational topic classification.  We also have studied the cross-lingual knowledge transfer from our Russian dataset to 50 different languages on parallel conversational data from the \texttt{MASSIVE} dataset.

\section{Related Work}\label{rel}

The community has proposed plenty of topical datasets. However, not all of these datasets are well-suited for conversational tasks. The majority of topical datasets consist of large pieces of written text (mostly -- news). Training on these datasets makes models overfit on the long pieces of data, which can lead to poor performance on conversational utterances. Moreover, the class nomenclature of these datasets is usually quite small -- therefore, a vast majority of topics one can bring up in the conversation are still out of their coverage. Furthermore, these datasets rarely contain Russian utterances.
Among the first examples of such datasets, we can mention \texttt{AG-NEWS}~\cite{ag_news_and_yahoo}, which has only four topics. We can also reference the dataset from \texttt{The Guardian}~\cite{guardian_authorship}. These datasets are also English-only. 

The news dataset \texttt{MLSUM}~\cite{mlsum} has versions for several languages (French, German, Turkish, Spanish, Russian). Due to the large size of news articles (compared to the conversational utterances), the examples in this dataset are too large for conversational tasks. Moreover, the 16 Russian topical classes from this dataset are derived from the news categories. These classes still don't cover a vast majority of conversational topics. 

The same problem of text's length also holds for the \texttt{XGLUE-nc}~\cite{xglue} dataset. This 10-class news dataset has an English-only training set, and a test sample from five European languages, including Russian.
An ontology dataset \texttt{DBpedia}~\cite{dbpedia} also suffers from this issue as it contains very long texts. Moreover, the nomenclature of this dataset (14 classes) is by no means sufficient for topical classification. 

Other topical datasets are too domain-specific, and thus they poorly fit for general-purpose tasks. Among such datasets, we can mention \texttt{LexGLUE}~\cite{lexglue} and \texttt{LEXTREME}~\cite{lextreme} benchmarks, which are focused on legal-specific topics. Other datasets are created for patent classification~\cite{hupd} and book title classification~\cite{blbooksgenre}. Russian datasets that were created for the classification of reviews on Russian medical facilities~\cite{healthcare_facilities_reviews} or classification of university-specific intents~\cite{pstu} can also be included in this category. However, the majority of conversational datasets are also very domain-specific, for example, conversational \texttt{NegoChat} dataset for negotiation domain~\cite{negochat}.

We can also mention the product review dataset from Amazon~\cite{amazon_reviews}. This dataset contains reviews of products sold on Amazon from different categories, grouped by the topic. However, the topics provided in this dataset are also insufficient for building a general-purpose topic classifier, as the possible range of topics to discuss differs from the variety of Amazon product categories. Additionally, the dataset does not support the Russian language.

One may also find interesting the idea of creating a topical dataset on the base of a question-answering website. Creators of the \texttt{Yahoo!Answers} dataset~\cite{ag_news_and_yahoo} have given a start to this idea. This 10-topic dataset contains questions and answers for topics from the "Yahoo Answers" service. However, the variety of topics included in this dataset is far from exhaustive. This dataset also does not contain the Russian language.

\texttt{MASSIVE}~\cite{massive} dataset is created for conversational topic and intent classification. In this dataset with ~17k samples (train+test+valid), one of the 18 topic classes and one of the 60 intent classes is assigned to every utterance. This dataset is massively multilingual, as every utterance in this dataset is provided in 51 different languages (including Russian), adapted to the specifics of corresponding countries. We can note that this dataset consists of conversational requests to a voice assistant. However, the nomenclature of topics provided in this dataset does not even remotely cover all possible user topics.

The nomenclature of covered topics in the dataset \texttt{DeepPavlov Topics}~\cite{dp_topics} is much larger, as 33 classes from this dataset cover a substantial number of possible conversational situations. However, this dataset does not comprise the Russian language.

The only publicly available Russian-language dataset we know that includes a significant number of conversational classes is \texttt{Chatbot-ru}~\cite{chatbotru}. This dataset has a very large nomenclature of Russian intents and topics (79 classes). However, the size of the dataset is too small for such amount of classes ($\sim$7.1k total samples). In this dataset, intents are treated in the same way as topics, so the real number of topical classes and samples in this dataset is smaller. Given that this dataset is also imbalanced, a vast majority of topics in this dataset have less than 100 samples per class (or even much less, up to 10-20).  Such a small number of samples per class makes the dataset suitable for the few-shot setting. However, it still leaves much room for improvement in terms of the dataset size expansion. Moreover, the variety of topical classes in this dataset is still incomplete and does not comprise some topics from~\cite{dp_topics}. 


As one can see from our review, not all topic datasets are suitable for use in a dialog system that works with real user phrases. Some datasets have too few classes, some other datasets have very domain-specific class nomenclature, and other datasets' examples are too different from the real-world dialog data which can cause additional distortions. Furthermore, the body of work in this field especially lacks topic datasets in Russian, as existing Russian datasets are incomplete and either too small or too specific.

The knowledge transfer from the Russian language for topic datasets is also under-researched. Our work aims to bridge this gap. 

\section{\texttt{RuQTopics} Dataset} 
This article examines the \texttt{RuQTopics} - Russian topic classification dataset. The raw data for this dataset were obtained from the \enquote{Yandex Que} question-answering service raw data.\footnote{\url{https://huggingface.co/datasets/its5Q/yandex-q/blob/main/full.jsonl.gz}}

The utterances in this dataset have 76 topics. We have selected the topics to utilize based on the dataset~\cite{dp_topics}. All utterances from this dataset contain questions. The questions in this dataset are short: 50\% of the questions have less than 10 words and less than 1\% - more than 30 words. At the same time, answers in this dataset are mostly very long: only ~1\% of the answers have less than 10 words, and 50\% of the answers have 65 words or less. 91.6\% of the answers consist of 256 words or less.

The topic of every question corresponded to its section on \enquote{Yandex Que}. For every question, we have selected the answer with the best quality score (or the first such answer, if there were several ones). For some questions, the answer was empty.


 We have split the question-answer pairs we obtained into two parts. In part 1 (single-label) we select only those pairs in which the question belongs to only one topic, and the answer to this question either does not exist or can be found solely in this topic. All other examples belong to part 2 (multi-label). Here and further, we work only with the single-label part of the \texttt{RuQTopics}.

 For all 76 topics, 532,590 unique questions were obtained, of which 403,938 are answered. The single-label part of the dataset contains 361,650 questions, of which 266,597 are answered. The multi-label part of the dataset contains 170,930 questions, of which 137,431 are answered. 

 Additionally, we have selected the matched part of the \texttt{RuQTopics} as a subset of the single-label one. If the question is answered, and the answer to this question can be found in only one topic (the same topic as the question has), the question-answer pair was included not only in the singlelabel part of the dataset but also in the matched part. 
 
Sizes of all parts of the \texttt{RuQTopics} for any class we use in this article can be found in Table~\ref{tab:data_sizes}. 



\begin{table}[t]
\caption{\texttt{RuQTopics} sizes for different splits and all classes considered in this article.}
\label{tab:data_sizes}
\centering
\scalebox{0.8}{
\begin{tabular}{|c|c|c|c|c|c|} \hline
\multirow{2}{*}{\textbf{data type}}  & \multicolumn{2}{c|}{\textbf{single-label}} & \multicolumn{2}{c|}{\textbf{multi-label}} & \multirow{2}{*}{\textbf{matched}} \\
\cline{2-5}
 & \multicolumn{1}{c|}{all} & \multicolumn{1}{c|}{answered} & \multicolumn{1}{c|}{all} & \multicolumn{1}{c|}{answered} & \\\hline
{Full dataset size} & 361,650 & 266,597 & 170,930 & 137,341 & 264,786\\
{6-class subset size} & 18864 & 15912 & 27191 & 20569 & 15830 \\ \hline
\textit{music} & 9,514 & 5,809 & 4,456 & 3,287 & 5,797\\ 
\textit{food, drinks and cooking} & 5,750 & 4,758 & 14,096 & 11,084 & 4,723\\ 
\textit{media and communications} & 4,505 & 2,637 & 5,577 & 3,948 & 2,619\\ 
\textit{transport} & 2,435 & 1,625 & 1,933 & 1,387 & 1,613\\
\textit{news} & 945 & 602 & 912 & 720 & 600\\ 
\textit{weather} & 890 & 481 & 217 & 143 & 478\\ \hline
\end{tabular}
}
\end{table}

We note that, as some \texttt{RuQTopics} classes are similar to each other, the applied utilization of this dataset might require merging some classes.

For our experiments on this dataset, we have trained the Transformer-based models with the hyperparameters and backbones described in the next section.
\section{Experimental Setup}
While training all models described in this work, we used the following hyperparameters: batch size 160, optimizer AdamW~\cite{adam}, betas (0.9,0.99), initial learning rate 2e-5, learning rate drops by 2 times if accuracy does not improve for 2 epochs, validation patience 3 epochs, max 100 training epochs. The max sequence length is 256 tokens. We performed three random restarts for all experiments and averaged the metrics.

We performed the experiments on multiple backbones from HuggingFace \texttt{Transformers} library~\cite{huggingface_transformers}, which all have similar BERT-like architecture:  \textit{bert-base-multilingual-cased}~\cite{multilingual_bert}, \textit{DeepPavlov/distilrubert-tiny-cased-conversational}~\cite{alina}, \textit{ai-forever/ruBert-base}~\cite{sbert_base} and \textit{DeepPavlov/rubert-base-conversational-cased}~\cite{rubert}. The models \textit{ai-forever/ruBert-base} and \textit{DeepPavlov/rubert-base-conversational-cased} are similar, but they have a slightly different number of parameters because of different tokenization. We describe the difference between these backbones in Table~\ref{tab:backbones}.

\subsection{Model Benchmarking}

To benchmark the performance of models trained on our dataset on the conversational tasks, we utilized the \texttt{MASSIVE} dataset for evaluation. We have selected this dataset because it contains data that were checked by the crowd workers, and it consists of the conversational utterances as well as \texttt{RuQTopics}.

While comparing our dataset with the \texttt{MASSIVE}, we saw that only six \texttt{MASSIVE} classes can be directly mapped to the \texttt{RuQTopics}. Therefore, we trained all described models only on the six corresponding classes from the single-label subset of \texttt{RuQTopics}: \textit{food, drinks, and cooking}  (corresponds to the \textit{cooking} \texttt{MASSIVE} class), \textit{news} (corresponds to the \textit{news} \texttt{MASSIVE} class), \textit{transport} (corresponds to the \textit{transport} \texttt{MASSIVE} class), \textit{music} (corresponds to the \textit{music} \texttt{MASSIVE} class), \textit{media and communication} (corresponds to the \textit{social} \texttt{MASSIVE} class) and \textit{weather} (corresponds to the \textit{weather} \texttt{MASSIVE} class). We did not merge \texttt{RuQTopics} classes even though it could have additionally improved the results for \textit{cooking} and \textit{transport} \texttt{MASSIVE} classes.

We validated all models on the Russian \texttt{MASSIVE} validation 6-class subset and tested them on the concatenation of train and test 6-class subsets of \texttt{MASSIVE}. Here and further, we denote this subset concatenation as a "custom test set".

This method allows testing the suitability of the dataset for conversational topic classification -- at least on a subset of classes. However, as examples for all classes were collected similarly, we expect that other classes from the \texttt{RuQTopics} are as suitable for the conversational topic classification as these six ones.

\section{Dataset Preprocessing}\label{prepr} 

\begin{table*}
\caption{Parameters of different backbone models considered in this article.}
\label{tab:backbones}
\centering
\scalebox{0.8}{
\begin{tabular}{|c|c|c|c|c|}
\hline
\multirow{1}{*}{\textbf{Backbone model}} & \multirow{1}{*}{\textbf{Abbreviation}}  & \textbf{Multilingual} &  \multicolumn{1}{c|}{\textbf{Layers}} & \textbf{Parameters}\\ 
\hline
\textit{DeepPavlov/distilrubert-tiny-cased-conversational}~\cite{alina} & \textit{rubert-tiny} & no & 2 & 107M\\ 
\textit{DeepPavlov/rubert-base-cased-conversational}~\cite{rubert} & \textit{rubert} & no & 12 & 177.9M\\ 
\textit{bert-base-multilingual-cased}~\cite{multilingual_bert} & \textit{multbert} & yes & 12 & 177.9M \\ 
\textit{ai-forever/ruBert-base}~\cite{sbert_base} & \textit{ru-sbert} & no & 12 & 178.3M\\ 
\hline
\end{tabular}
}
\end{table*}

We needed to identify the best method of \texttt{RuQTopics} preprocessing for the best performance on conversational tasks. Specifically, we have compared five different methods of preprocessing the \texttt{RuQTopics} dataset. We name them as "modes" in Table~\ref{tab:matched}. In these modes:
\begin{itemize}
    \item \textbf{Q} means using only questions.
    \item \textbf{A} means using only answers.
    \item \textbf{Q [SEP] A} means using the concatenation of every question with the corresponding answer using [SEP] token. If the question is unanswered, it means using only a question.
\end{itemize}

For all of these preprocessing methods, we performed training on the matched version of the \texttt{RuQTopics} (column "matched" from Table~\ref{tab:data_sizes}). This training mode allows making the apple-to-apple comparison between features obtained by different preprocessing methods, as the number of training samples in this method is the same regardless of how we preprocess the data. We present in Table~\ref{tab:matched} the results obtained in this training mode. We also present in Table~\ref{tab:full} the results obtained by training on the full single-label version of this dataset ( column "singlelabel" from Table~\ref{tab:data_sizes}).

\begin{table*}
\centering
\caption{Accuracy (F1) of different kinds of backbones on the custom test set of Russian \texttt{MASSIVE}. The models were trained on the \texttt{RuQTopics} 6-class \textbf{matched} subsets preprocessed using different preprocessing modes described in Section~\ref{prepr}. We selected these six classes as they could be mapped on the MASSIVE dataset. Backbones are abbreviated as in Table~\ref{tab:backbones}. Averaged by three runs.}
\scalebox{1}{
\label{tab:matched}
\begin{tabular}{|c|c||c|c|c|c|c|c|c|c|c|c|c|c|c|c|}
\hline
\multirow{2}{*}{\textbf{Model}} & \multirow{2}{*}{\textbf{Mode}}  &  \multicolumn{2}{c|}{\textbf{Total}} &  \multicolumn{2}{c|}{\textbf{music}} &  \multicolumn{2}{c|}{\textbf{cooking}} &  \multicolumn{2}{c|}{\textbf{news}} &  \multicolumn{2}{c|}{\textbf{transport}} &  \multicolumn{2}{c|}{\textbf{weather}} &  \multicolumn{2}{c|}{\textbf{social}} \\ 
\cline{3-16}
& & Acc & Mc-F1 & Acc & F1 & Acc & F1 & Acc & F1 & Acc & F1 & Acc & F1 & Acc & F1 \\ \hline
\textit{ru} &  \textbf{Q} & 84.2 & 83.4 & 94.3 & 87.5 & 99.0 & 82.0 & 80.1 & 83.1 & 92.3 & 90.7 & 81.3 & 89.0 & 65.5 & 68.1\\ 
\textit{rutiny} &  \textbf{Q} & 85.7 & 84.9 & 94.3 & 89.1 & 99.2 & 81.5 & 76.8 & 83.3 & 93.8 & 92.4 & 84.2 & 90.4 & 73.2 & 72.7\\ 
\textit{rusber} &  \textbf{Q} & 85.2 & 84.1 & 93.9 & 90.2 & 98.9 & 79.8 & 80.3 & 84.3 & 92.8 & 90.6 & 85.7 & 90.9 & 64.9 & 69.0\\ 
\textit{mult} &  \textbf{Q} & 79.1 & 77.7 & 93.2 & 79.8 & 98.0 & 74.0 & 73.5 & 80.3 & 89.9 & 87.9 & 74.4 & 83.9 & 55.2 & 60.1\\ \hline
\textit{ru} &  \textbf{A} & 80.7 & 79.1 & 95.8 & 78.5 & 98.2 & 81.6 & 66.3 & 77.2 & 91.5 & 87.9 & 87.1 & 91.1 & 51.6 & 58.3\\ 
\textit{rutiny} &  \textbf{A} & 82.4 & 81.0 & 96.4 & 81.1 & 99.3 & 78.1 & 71.3 & 80.8 & 88.9 & 90.3 & 86.7 & 90.8 & 60.0 & 64.8\\ 
\textit{rusber} &  \textbf{A} & 82.3 & 80.7 & 94.9 & 81.2 & 98.9 & 80.6 & 72.6 & 80.8 & 89.8 & 89.2 & 88.7 & 91.2 & 54.5 & 61.3\\ 
\textit{mult} &  \textbf{A} & 76.8 & 75.3 & 94.3 & 76.6 & 96.1 & 70.0 & 68.3 & 77.5 & 83.0 & 83.4 & 78.6 & 85.2 & 50.8 & 59.0\\ \hline
\textit{ru} &  \textbf{Q [SEP] A} & 85.7 & 85.2 & 92.3 & 90.1 & 97.4 & 86.4 & 79.1 & 82.9 & 93.3 & 91.5 & 86.4 & 91.1 & 70.3 & 69.5\\ 
\textit{rutiny} &  \textbf{Q [SEP] A} & 85.0 & 84.2 & 95.3 & 87.2 & 98.3 & 82.4 & 75.5 & 82.3 & 89.7 & 92.0 & 86.4 & 91.1 & 72.4 & 70.5\\ 
\textit{rusber} &  \textbf{Q [SEP] A} & 85.3 & 84.7 & 92.7 & 89.5 & 98.7 & 85.8 & 80.7 & 82.1 & 91.0 & 91.1 & 87.1 & 92.0 & 66.7 & 67.9\\ 
\textit{mult} &  \textbf{Q [SEP] A} & 78.5 & 77.6 & 93.0 & 82.5 & 95.9 & 72.9 & 67.0 & 77.1 & 86.1 & 85.6 & 75.1 & 84.0 & 65.1 & 63.5\\ \hline
\end{tabular}
}
\end{table*}

\begin{table*}
\centering
\caption{Accuracy (F1) of different kinds of backbones on the custom test set of Russian \texttt{MASSIVE}. The models were trained on the \texttt{RuQTopics} 6-class \textbf{full} subsets preprocessed using different preprocessing modes described in Section~\ref{prepr}. We selected these six classes as they could be mapped on the MASSIVE dataset. Backbones are abbreviated as in Table~\ref{tab:backbones}. Averaged by three runs.}
\scalebox{1}{
\label{tab:full}
\begin{tabular}{|c|c||c|c|c|c|c|c|c|c|c|c|c|c|c|c|}
\hline
\multirow{2}{*}{\textbf{Model}} & \multirow{2}{*}{\textbf{Mode}}  &  \multicolumn{2}{c|}{\textbf{Total}} &  \multicolumn{2}{c|}{\textbf{music}} &  \multicolumn{2}{c|}{\textbf{cooking}} &  \multicolumn{2}{c|}{\textbf{news}} &  \multicolumn{2}{c|}{\textbf{transport}} &  \multicolumn{2}{c|}{\textbf{weather}} &  \multicolumn{2}{c|}{\textbf{social}} \\ 
\cline{3-16}
& & Acc & Mc-F1 & Acc & F1 & Acc & F1 & Acc & F1 & Acc & F1 & Acc & F1 & Acc & F1 \\ \hline
\textit{ru} &  \textbf{Q} & 85.0 & 84.3 & 94.7 & 87.5 & 98.4 & 86.0 & 82.5 & 82.7 & 92.1 & 92.1 & 82.5 & 89.6 & 66.1 & 68.0\\ 
\textit{rutiny} &  \textbf{Q} & 85.7 & 85.2 & 95.0 & 87.5 & 98.7 & 87.3 & 82.2 & 82.9 & 92.8 & 92.3 & 84.3 & 90.6 & 67.3 & 70.3\\ 
\textit{rusber} &  \textbf{Q} & 85.5 & 84.9 & 93.7 & 89.9 & 98.8 & 87.2 & 83.0 & 83.1 & 93.1 & 91.1 & 85.1 & 91.1 & 64.3 & 67.2\\ 
\textit{mult} &  \textbf{Q} & 80.8 & 79.8 & 94.3 & 77.3 & 97.2 & 82.8 & 75.5 & 81.1 & 90.0 & 90.5 & 78.5 & 86.0 & 57.5 & 61.0\\ \hline
\textit{ru} &  \textbf{Q [SEP] A} & 85.4 & 84.9 & 94.0 & 88.5 & 97.6 & 87.1 & 82.5 & 83.7 & 93.1 & 91.5 & 83.6 & 90.0 & 67.1 & 68.8\\ 
\textit{rutiny} &  \textbf{Q [SEP] A} & 85.3 & 84.7 & 94.3 & 87.4 & 97.9 & 86.4 & 79.3 & 81.4 & 91.6 & 92.8 & 86.0 & 91.0 & 68.2 & 69.1\\ 
\textit{rusber} &  \textbf{Q [SEP] A} & 85.1 & 84.2 & 93.1 & 91.6 & 98.3 & 88.4 & 88.1 & 81.2 & 93.3 & 91.9 & 86.2 & 91.6 & 53.7 & 60.7\\ 
\textit{mult} &  \textbf{Q [SEP] A} & 80.0 & 79.7 & 94.2 & 77.0 & 95.1 & 85.5 & 74.9 & 80.3 & 87.8 & 88.0 & 73.8 & 83.8 & 64.6 & 63.6\\ \hline
\end{tabular}
}
\end{table*}
As one can see from Table~\ref{tab:matched}, the question-only setting yields larger scores than the answer-only setting. This conclusion holds for all considered backbone models, proving that the questions are the most informative feature in the \texttt{RuQTopics} dataset.
If we concatenate questions to answers, the scores do not change significantly compared to the question-only setting.

We have also tried using answers that are summarized by TextRank~\cite{summarizer} instead of the full answers in the experiments. The summarized answer-only setting has shown sustainably worse results than the answer-only one, and the concatenation of questions to summarized answers has given the same scores as the concatenation of questions to answers.

Overall, all Russian models show similar results, and the multilingual model expectedly trails behind them all.

All these conclusions are also valid for the full 6-class subset, as one can see from Table~\ref{tab:full}. 
For the experiments in the next sections, we chose the \textbf{Q} preprocessing mode, as all other modes are either more complicated and give no better results (\textbf{Q [SEP] A}), or show worse results (\textbf{A}).

\section{Evaluation for all \texttt{RuQTopics} classes}

Another important task is to figure out how well the \texttt{RuQTopics} classes can be distinguished from each other. To do so, we perform 5-fold cross-validation on all questions from the singlelabel \texttt{RuQTopics} part. We present the results in Table~\ref{tab:crossvalidation}.

\begin{table*}
\centering
\caption{Accuracy (Macro-F1) of different backbone models for the 5-fold cross-validation on all questions from the singlelabel part of the \texttt{RuQTopics} dataset (76 classes). Backbones are abbreviated as in Table~\ref{tab:backbones}.}
\scalebox{1}{
\label{tab:crossvalidation}
\begin{tabular}{|c||c|c|c|c|c|c|c|c|c|c|c|c|}
\hline
\multirow{2}{*}{\textbf{Model}} & \multicolumn{2}{c|}{\textbf{Average}} & \multicolumn{2}{c|}{\textbf{Fold 1}} & \multicolumn{2}{c|}{\textbf{Fold 2}} &\multicolumn{2}{c|}{\textbf{Fold 3}} & \multicolumn{2}{c|}{\textbf{Fold 4}} & \multicolumn{2}{c|}{\textbf{Fold 5}} \\ 
\cline{2-13}
& Acc & Mc-F1 & Acc & Mc-F1 &  Acc & Mc-F1 & Acc & Mc-F1 & Acc & Mc-F1 &  Acc & Mc-F1 \\ \hline
\textit{rusber} & 74.0 & 53.4 & 73.7 & 54.3 & 73.8 & 52.8 & 73.9 & 53.0 & 74.1 & 54.2 & 74.2 & 52.9\\
\textit{ru} & 73.7 & 52.5 & 73.5 & 52.9 & 73.7 & 51.9 & 73.6 & 52.3 & 73.9 & 53.1 & 73.9 & 52.3\\
\textit{rutiny} & 72.2 & 50.9 & 72.0 & 49.7 & 72.2 & 50.9 & 72.0 & 51.4 & 72.4 & 51.1 & 72.3 & 51.6\\
\textit{mult} & 71.4 & 51.9 & 71.2 & 52.4 & 71.5 & 51.9 & 71.5 & 51.4 & 71.2 & 51.6 & 71.7 & 52.1\\ \hline
\end{tabular}
}
\end{table*}

The results could have been additionally improved by merging some classes from similar \enquote{Yandex. Que} topics. But even without that, Russian non-distilled backbones show an accuracy of 73.7-74.0\%, whereas the Russian distilled backbones fares slightly worse (72.2\% accuracy). The multilingual backbone expectably trails slightly behind these backbones by this measure (71.4\% accuracy). This shows that the topical classes in the dataset can be distinguished from each other with sufficiently high accuracy.

\section{Cross-Lingual Knowledge Transfer} 

After we had selected the best setting, the following questions emerged: how effectively does the knowledge from this setting transfer across multiple languages? And what influences the efficiency of this transfer? To answer these questions, we pre-trained \textit{bert-base-multilingual-cased}, which allows effective cross-lingual transfer learning on different NLP tasks~\cite{ner,squad}, on the data from \textbf{full} validation 6-class \texttt{RuQTopics} subset, which are preprocessed by the \textbf{Q} preprocessing mode. For this backbone, using the full subset instead of the matched subset gave ~1-2\% growth in accuracy and macro-F1 for the Russian language. 

In this stage, we infer this model not only on the Russian \texttt{MASSIVE} but also on all other languages it contains.\footnote{We use \texttt{MASSIVE} version 1.1, which contains the Catalan language. For the Chinese language, we have utilized both sets of characters as \texttt{MASSIVE} has two Chinese versions.}

An interesting research question is the correlation of the model quality for different languages with the pretraining sample size for that language. The authors of the \textit{bert-base-multilingual-cased} claim~\cite{multilingual_bert} that the learning sample for every utilized language was comprised of the Wikipedia texts for that language and that they performed an exponential smoothing of the training sample with the factor of 0.7 to balance the languages. Therefore, as a proxy of the Wikipedia size for every language, we used the number of articles in the Wikipedia of this language at the time of the BERT article's release, smoothed by the factor of 0.7. 

We present the metrics obtained by the evaluation of the multilingual BERT on the custom \texttt{MASSIVE} test subset for all languages in Table~\ref{crosslingual}. For every language, we also provide the genealogical distance to Russian (calculated as in~\cite{lang_sim}) and the original Wikipedia size we used in the same table.

\begin{table}[!htbp]
\caption{Accuracy (F1) of the \textit{bert-base-multilingual-cased} on the custom test set for all \texttt{MASSIVE} languages. The model was trained on the \textbf{Q} version of \textbf{full} \texttt{RuQTopics} 6-class subset and validated on the 6-class validation set of Russian \texttt{MASSIVE}. \textbf{Code} means ISO 639-1 language code, \textbf{Dist} means genealogical distance between that language and Russian~\cite{lang_sim}. \textbf{N} means the number of Wikipedia articles in that language as of 11-10-2018. We trained on the \textbf{full} single-label version of \texttt{RuQTopics}. Averaged by three runs.}
\label{crosslingual}
\begin{minipage}{0.5\textwidth}
\scalebox{0.85}{
\begin{tabular}[baseline={(0,2.1)}]{|c|c|c|c|c|c|} \hline
\multirow{2}{*}{\textbf{Language}}  & \multirow{2}{*}{\textbf{Code}} & \multirow{2}{*}{\textbf{Dist}} & \multirow{2}{*}{\textbf{N}}  &  \multicolumn{2}{c|}{\textbf{Metrics}} \\ 
\cline{5-6}
& & & & Acc & Mc-F1 \\ \hline
Russian & ru & 0 & 1,501,878 & 80.8 & 79.8\\
Chinese-TW & zh-TW & 92.2 & 1,025,366 & 79.6 & 79.1\\
Chinese & zh & 92.2 & 1,025,366 & 78.0 & 77.7\\
English & en & 60.3 & 5,731,625 & 75.2 & 75.6\\
Japanese & ja & 93.3 & 1,124,097 & 72.4 & 70.5\\
Slovenian & sl & 4.2 & 162,453 & 70.3 & 69.0\\
Swedish & sv & 59.5 & 3,763,579 & 70.2 & 69.6\\
Malay & ms & n/c & 320,631 & 68.9 & 67.7\\
Italian & it & 45.8 & 1,466,064 & 68.8 & 68.0\\
Indonesian & id & 91.2 & 440,952 & 68.7 & 67.5\\
Dutch & nl & 64.6 & 1,944,129 & 68.7 & 68.5\\
Portuguese & pt & 61.6 & 1,007,323 & 68.6 & 68.7\\
Spanish & es & 51.7 & 1,480,965 & 68.2 & 68.0\\
Danish & da & 66.2 & 240,436 & 67.8 & 66.7\\
French & fr & 61.0 & 2,046,793 & 65.5 & 65.5\\
Persian & fa & 72.4 & 643,750 & 65.2 & 64.2\\
Turkish & tr & 86.2 & 316,969 & 64.5 & 62.4\\
Vietnamese & vi & 95.0 & 1,190,187 & 64.3 & 65.1\\
Norwegian B & nb & 67.2 & 495,395 & 64.3 & 64.0\\
Polish & pl & 5.1 & 1,303,297 & 64.2 & 62.2\\
Azerbaijani & az & 87.7 & 138,538 & 63.9 & 63.1\\
Catalan & ca & 60.3 & 591,783 & 61.4 & 60.4\\
Hungarian & hu & 87.2 & 437,984 & 61.3 & 60.0\\
Hebrew & he & 88.9 & 231,868 & 60.9 & 59.5\\
Hindi & hi & 69.8 & 127,044 & 60.7 & 58.7\\
\hline
\end{tabular}
}
\end{minipage}
\begin{minipage}{0.5\textwidth}
\scalebox{0.85}{
\begin{tabular}[baseline={(0,2.1)}]{|c|c|c|c|c|c|} \hline
\multirow{2}{*}{\textbf{Language}}  & \multirow{2}{*}{\textbf{Code}} & \multirow{2}{*}{\textbf{Dist}} & \multirow{2}{*}{\textbf{N}}  &  \multicolumn{2}{c|}{\textbf{Metrics}} \\ 
\cline{5-6}
& & & & Acc & Mc-F1 \\ \hline
Korean & ko & 89.5 & 429,369 & 60.4 & 59.6\\
Romanian & ro & 55.0 & 388,896 & 57.1 & 53.9\\
Urdu & ur & 66.7 & 140,939 & 56.4 & 55.9\\
Arabic & ar & 86.5 & 619,692 & 56.2 & 55.7\\
Kannada & kn & 90.8 & 23,844 & 56.1 & 53.0\\
Filipino & tl & 91.9 & 80,992 & 55.0 & 51.3\\
Telugu & te & 96.7 & 69,354 & 53.7 & 49.3\\
Finnish & fi & 88.9 & 445,606 & 53.3 & 51.3\\
Burmese & my & 86.0 & 39,823 & 52.5 & 49.7\\
Afrikaans & af & 64.8 & 62,963 & 52.4 & 50.3\\
Tamil & ta & 94.7 & 118,119 & 52.4 & 50.1\\
German & de & 64.5 & 2,227,483 & 52.2 & 51.6\\
Albanian & sq & 69.4 & 74,871 & 51.5 & 47.2\\
Latvian & lv & 49.1 & 88,189 & 49.6 & 48.4\\
Malayalam & ml & 96.7 & 59,305 & 48.7 & 46.3\\
Armenian & hy & 77.8 & 246,571 & 48.1 & 47.5\\
Bangla & bn & 66.3 & 61,294 & 47.3 & 45.3\\
Thai & th & 89.5 & 127,010 & 46.5 & 44.9\\
Greek & el & 75.3 & 153,855 & 46.3 & 44.8\\
Georgian & ka & 96.0 & 124,694 & 39.2 & 38.1\\
Javanese & jv & 95.4 & 54,964 & 38.7 & 37.1\\
Mongolian & mn & 86.2 & 18,353 & 36.6 & 33.7\\
Icelandic & is & 68.9 & 45,873 & 32.6 & 29.9\\
Swahili & sw & 95.1 & 45,806 & 31.0 & 28.0\\
Welsh & cy & 75.5 & 101,472 & 28.5 & 25.3\\
Khmer & km & 97.1 & 6,741 & 16.1 & 8.6\\
Amharic & am & 86.6 & 14,375 & 12.1 & 5.0\\
\hline
\end{tabular}
}
\end{minipage}
\end{table}

The Spearman correlation of the total accuracy with the smoothed Wikipedia size is 0.773 (p-value 2.997e-11, 95\% CI: [0.63, 0.86]). At the same time, the Spearman correlation of the total accuracy with the genealogical distance to the Russian is -0.323 (p-value 0.022, 95\% CI: [-0.55, -0.05]).\footnote{We excluded the Russian language itself from the calculations.}  If we take into account the smoothed Wikipedia size as the confounding variable, the partial correlation of the total accuracy with the genealogical distance to the Russian becomes -0.027 (p-value 0.856, 95\% CI: [-0.31, 0.26]), which is statistically insignificant.

\section{Discussion} 
As one can see, the \texttt{RuQTopics} dataset overall suits fairly well for the conversational topic classification.

We suppose that, apart from the topical classification, this dataset can also be utilized for the question-answering task. However, we leave the checking of this statement for future research.

In the case of question classification, different Russian-only baseline models trained on the \texttt{RuQTopics} 6-class subset obtain an accuracy of around 85\% on the subset of the same six classes from the Russian \texttt{MASSIVE} (Table~\ref{tab:full}). 

We obtain such accuracy only if we utilize questions from \texttt{RuQTopics} in the training features (either by themselves or in concatenation with answers), which proves that the questions are the most informative features in this dataset. 

Surprisingly, switching between different Russian-only baseline models, including even the two-layer distilled one, did not significantly alter the results. That proves that the distilled conversational models suit well for conversational tasks, especially in the case of constrained computational resources.

For training models on all 76 classes of the \texttt{RuQTopics} in the question-only setting, all backbones show accuracy above 70\%. That shows that the dataset is suitable for the classification task as a whole, not just as a six-class subset.

In the case of evaluation of the multilingual BERT (trained on the \texttt{RuQTopics} question subset) on all languages included in the \texttt{MASSIVE} dataset, the accuracy by language closely correlates with the approximated size of the BERT pretraining dataset for that language (Spearman correlation 0.773 with p-value 2.997e-11). We have approximated the dataset size by exponentiation of the language-wise number of Wikipedia size as of 11-10-2018 (date of release of the~\cite{bert}) by 0.7, analogously to the original article. 

 Such correlation was obtained even though an average Wikipedia article in different languages has a different number of tokens and sentences. We suppose that if we had the precise number of training samples for every language that the original multilingual model received at the pretraining stage, the correlation would have been even higher; but the authors of the original BERT article provided neither the original training sample nor its language-wise size. 

 At the same time, the correlation of the model scores with the genealogical distance to the Russian is statistically insignificant. This leads to the conclusion that the main factor determining the quality of knowledge transfer between different languages in the multilingual BERT-like models is, by far, the size of the pretraining sample for this language. We can suppose that for the case of languages that are very linguistically close (e.g. Russian and Belarusian) such closeness also impacts knowledge transfer, but examining the importance of this factor requires additional research. 

\section{Conclusion} 
This article investigates the knowledge transfer from the \texttt{RuQTopics} dataset. This Russian topical dataset combines a large sample number (361,560 single-label, 170,930 multi-label) with extensive class coverage (76 classes). We have prepared this dataset from the \enquote{Yandex Que} raw data.

By evaluating the \texttt{RuQTopics} -- trained models on the six matching classes of the Russian \texttt{MASSIVE} subset, we have proved that the \texttt{RuQTopics} dataset is suitable for real-world conversational tasks, as the Russian-only models trained on this dataset consistently yield the accuracy around 85\% on this subset (Table~\ref{tab:full}). We also have figured out that for the multilingual BERT, trained on the \texttt{RuQTopics} and evaluated on the same six classes of \texttt{MASSIVE} (for all \texttt{MASSIVE} languages), the language-wise accuracy closely correlates with the approximate size of the pretraining BERT's data for the corresponding language. At the same time, the correlation of the language-wise accuracy with the genealogical distance from the Russian is not statistically significant.

\section{Acknowledgments}
We express gratitude to Pavel Levchuk for the raw data collection, to Anastasiya Chizhikova for her help with the English language, and to Alexander Popov for valuable remarks.

\bibliographystyle{unsrtnat}

\end{document}